# Self-supervised Learning of Occlusion Aware Flow Guided 3D Geometry Perception with Adaptive Cross Weighted Loss from Monocular Videos

Fang Jiaojiao
School of Electronics and
Information Engineering
*Xi'an Jiaotong University*
*Xi'an, China 710049*
995541569@qq.com

Liu Guizhong[*]
School of Electronics and
Information Engineering
*Xi'an Jiaotong University*
*Xi'an, China 710049*
liugz@xjtu.edu.cn

*Abstract*—Self-supervised deep learning-based 3D scene understanding methods can overcome the difficulty of acquiring the densely labeled ground-truth and have made a lot of advances. However, occlusions and moving objects are still some of the major limitations. In this paper, we explore the learnable occlusion aware optical flow guided self-supervised learning of depth and ego-motion with an adaptive cross weighted loss to address the above limitations. Firstly, we try to learn the self-discovered occlusion mask fused optical flow network by an occlusion aware photometric loss combined with temporally supplemental information and backward-forward consistency between adjacent views. And then, we design an adaptive cross-weighted loss of the geometric and photometric error between the cross tasks to distinguish the moving objects that violate the static scene assumption. Our method shows promising results on KITTI, Make3D and Cityscapes datasets on multiple tasks, and also shows good generalization ability under a variety of challenging scenarios.

## I. INTRODUCTION

Understanding a scene's 3D geometric structure from video sequences is a key yet challenging task. It is widely applied in robotics, autonomous driving, scene interaction, and augmented reality. The vision-based 3D scene geometry understanding has been studied extensively. The traditional methods mainly rely on the structure-from-motion (SfM) frameworks that leverage sparse hand-crafted features, and the multiple view geometry under the re-projection error [5]. However, these methods could only offer sparse results. With the advent of deep learning, recently academics have tried to formulate a supervised deep convolutional neural network (CNN) based depth and pose learning framework. This produces satisfying results but suffers the inherent limitation of depending on large densely labeled training sets, which are difficult and laborious to obtain.

Recently, the self-supervised frameworks are proposed [6][7] to learn the depth and pose networks without ground-truth by minimizing photometric loss among adjacent views. These systems implicitly embed basic 3D geometry and achieved competitive performance, but are still affected by occlusions, moving objects, and face difficulties in challenging scenarios. The incorrect photometric loss would mislead the training process of all tasks. The MaskFlowNet [8] learns to distinguish the occlusions by a learnable mask without any explicit occlusion signal supervision in the supervised optical flow learning and achieves better performance. Chen *et al.* proposed [9] a wonderful idea to deal with the dynamic nature of the real-world scenarios in the self-supervised depth-pose learning framework by an adaptive photometric loss. Based on the brilliant thinking, we further consider the followings: 1) training data reduction for depth, ego-motion, and optical flow due to the minimum error based adaptive loss; 2) the occlusions, 3) the optical flow and camera pose-based epipolar geometric loss is sensitive to the moving objects.

In this work, we introduce an occlusion self-discovered optical flow-guided self-supervised depth and pose learning framework. We first explore to train the learnable occlusion mask combined optical flow network by the self-supervised manner with an occlusion-aware photometric loss. Then an adaptive cross-weighted loss of photometric and geometric error between the optical flow and the ego-motion and depth is used to exclude the dynamic objects in calculating the training loss. The main contributions in this work are as follows:

i) To deal with occlusions, we import the mask optical flow network and jointly learn the occlusion aware optical flow with depth-pose in a self-supervised manner and gain significant improvements on multiple tasks. Furthermore, we can obey the self-supervised nature of the existing depth, pose and optical flow learning framework, and automatically distinguish the occlusions for all tasks.

ii) The photometric loss of the depth and camera pose networks learning will be invalid in the motion object regions, while the optical flow based photometric loss is more robust in these situations. While the epipolar geometric constraints of these tasks are just the opposite in the motion areas. Thus we consider the photometric and geometric error simultaneously to construct an adaptive cross weighted loss between the depth-pose and optical flow which could enhance the robustness of the model against moving objects.

The widely evaluated experiments on the KITTI dataset show the effectiveness of our method over the state-of-the-arts without online refinement. The overview of the proposed method is shown in Fig. 1.

## II. RELATED WORKS

The vision-based 3D scene understanding encompasses many basic tasks such as depth estimation, optical flow, visual odometry, etc [9]. Traditional feature matching and structure-from-motion (SfM) based methods [4],[5] had a mature mathematical theory system and achieved remarkable results. But the reconstruction results were usually sparse and invalid in low-texture or occluded areas.

---

[*] *Corresponding author*

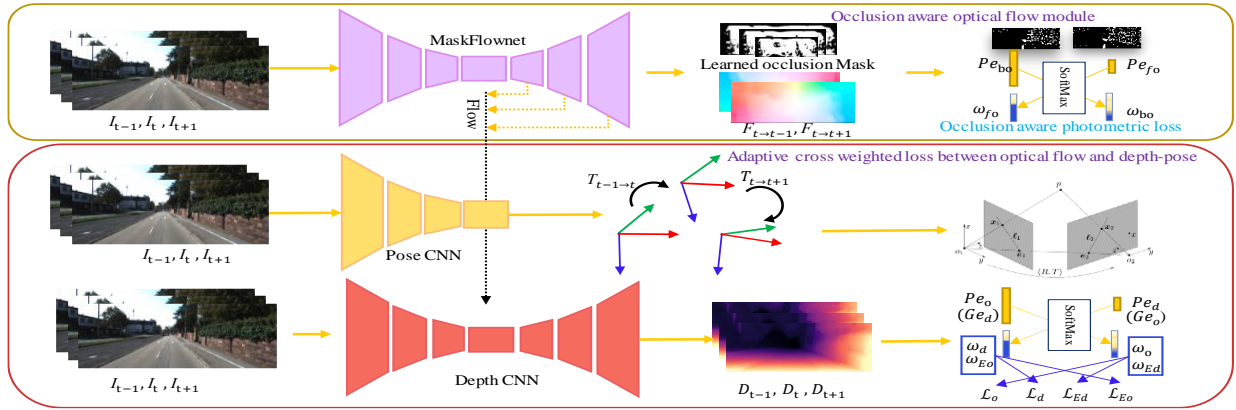

Fig. 1. An overview of the proposed framework. Our model imports the learnable occlusion mask optical flow into the self-supervised depth-pose learning framework. Then an adaptive cross weighted loss is used for the self-supervised learning of depth-pose and optical flow.

To settle the weakness of sparse hand-crafted features, recent methods introduced the deep learning based scene geometry understanding methods via a lot of densely labeled ground-truth data, such as monocular depth estimation [13],[14], optical flow [16],[17],[18] and camera pose estimation [20],[21]. These methods relied on expensive LIDAR equipment for data collection [1],[2] and formulated it as a pure regression problem that cares little about geometric relations. To make use of the multiple ingredients, the boundary detection [23], surface normal prediction [12],[15],[24], and semantic segmentation [26],[27] were fused together with depth learning to construct a jointly learning framework. To solve the data insufficiency problem, synthesis data was used to train the depth network [25],[28].

To weak the demand for densely labeled ground truth data during training, a large amount of self-supervised learning methods had emerged [7],[29]. The central idea was to construct a photometric error as indirect supervision based on a differentiable warping between adjacent views, which had been successfully applied to optical flow estimation [31],[32],[43]. Along this line, several methods [35],[38],[44] further improved the performance by incorporating additional constraints including Iterative Closest Point registration regularization [35], epipolar geometry [44], and collaborative competition [46], or better network structures [42]. And later, an online fine-tuning module was proposed to decrease the gap between the training and testing [9]. To gain the reliability of the depth and camera pose learning, some related tasks, such as the optical flow [33],[34],[46] and semantic segmentation [41],[45], were introduced into the self-supervised depth and pose learning framework to constrain the cross-task consistency. To solve the scale inconsistency and ambiguity problem, some recent methods introduced a multi-view geometry consistency [36],[37], or replaced the pose network with geometric-based methods[39],[40]. To handle occlusions, some works used the forward-backward optical flow consistency to exclude the occlusion areas from the photometric loss computation [33],[39],[46]. However, the occlusion estimation used by these methods required bidirectional optical flow estimation, which limited its flexibility. The MaskFlowNet [8] used deformable convolution and an autonomic learned occlusion mask to solve the invalid warping problem in the occlusion areas. To deal with the moving objects, recent works introduced some analytical masks [35],[47] and self-discovered masks [36],[44] to exclude these invalid regions.

This paper aims to integrate advantages of (a) excellent occlusion aware optical flow network; (b) the adaptive cross weighted loss of photometric error and geometric error between optical flow and the rigid scene flow to deal with the moving objects.

## III. THE PROPOSED METHODOLOGY

Our goal is to facilitate the self-supervised learning of depth and camera pose from monocular videos by the occlusion aware optical flow, and train them by a robust photometric and geometric combined loss. Given the unlabeled source frames $I_s$, where $s \in \{t-1, t+1\}$ and the target frame $I_t \in R^{H \times W \times 3}$ taken by a motion camera with an intrinsic $K$, we first predict the optical flow $F_{s \to t}$ between them using the MaskFlowNet. And then their depth maps $(D_t, D_s)$ are estimated by the depth network, the 6DoF camera pose $T_{s \to t} = [R_{s \to t}, t_{s \to t}]$ from the time $s$ to $t$ using the pose network.

### A. Self-supervised Learning of Occlusion Self-discovered Optical Flow

The MaskFlowNet [9] is proposed to self-discover the occlusions during the feature matching and achieved state-of-the-art performance in supervised optical flow learning. We will start with the structure of the network. Similar to the PWC-Net [18], multi-scales feature maps of the target view $I_t$ and source views $I_s$ are first extracted by a siamese convolutional network. Then, the extracted feature maps of the frame $I_s$ is aligned with the feature map of $I_t$ at each scale $l$ based on the upsampled flow estimation $F_{t \to s}^{l+1}$ at scale $l+1$. Then a cost volume is computed by a correlation layer, which is subsequently concatenated with the previous extracted feature map of $I_t$ and the upsampled optical flow at scale $l+1$. Finally, the outputs of the flow estimation layer is element-wise summation with the upsampled optical flow at scale $l+1$ to generate the optical flow at scale $l$. Iterating this process at all scale levels can predict the multi-scales optical flow.

To deal with the mismatching problem of the correlation layer that occurred at the occluded areas, a learnable occlusion-aware mask called occlusion-aware asymmetric feature matching module [8] is incorporated into the structure mentioned above. The warped feature maps $W \in R^{B \times H \times W \times C}$ are element-wise multiplied by the learnable occlusion mask $M_{occ} \in R^{B \times H \times W \times 1}$ and then added with an extra trade-off tensor $\mu \in R^{B \times H \times W \times C}$ to facilitate the learning process. $M_{occ}$ is normalized to be within the range of [0, 1]. While the warping operation is replaced by the deformable convolution to deal with the asymmetry of the feature matching process in the occluded areas. The learned occlusion mask can also be used to deal with the occlusions in self-supervised learning of monocular depth and camera pose.

In this paper, we try to train the occlusion discovered optical flow network in the self-supervised manner by an occlusion aware photometric loss which measures the difference between the actual collected image and the synthesized image. Referring to the commonly applied strategies to deal with occlusions, we train the optical flow network by combining the temporal supplemental information and the consistency between the optical flow from time $s$ to $t$ and time $s$ to $t$ its inverse flow to exclude the valid supervision information at occlusion area. Given the forward optical flow $F_{t \to s}$ and the backward optical flow $F_{s \to t}$, we can use the backward optical flow to generate an occlusion map $M_{t \to s} \in R^{B \times H \times W \times 1}$ by modeling the non-occluded area in $I_t$ as the range of $F_{s \to t}$, which can be calculated by

$$R(x, y) = \sum_{i=1}^{W} \sum_{j=1}^{H} \max(0, 1 - |x - (i + F_{s \to t}^{x}(i, j))|) \cdot \max(0, 1 - |y - (j + F_{s \to t}^{y}(i, j))|) \quad (1)$$

Where $R(x, y)$ indicates the number of correspondences between adjacent images at position $(x, y)$ [32], $(H, W)$ are height and width of optical flow. $F_{s \to t}^{x}$ and $F_{s \to t}^{y}$ are the components of $F_{s \to t}$ in the horizontal and vertical directions respectively. The occlusion map can be expressed as $M_{t \to s}(x, y) = \min(1, R(x, y))$. As the scene in the two moments $t-1$ and $t+1$ are usually complementary, we also utilize the adaptive loss between the photometric errors $L_{bo}$ and $L_{fo}$ of the two directions $t \to t-1$ and $t \to t+1$ to exclude occlusions. Instead of using the minimum error, here we use an adaptive weight computed by the Softmax function. Thus the total occlusion aware photometric loss for optical flow learning can be expressed as

$$L_{op} = \omega_{bo} M_{t \to t-1} L_{bo} + \omega_{fo} M_{t \to t+1} L_{fo} \quad (2)$$

Where $L_{bo}$ and $L_{fo}$ are the photometric losses defined as the convex combination of the L1 norm and SSIM. The adaptive per-pixel weights $\omega_{bo}$ and $\omega_{fo}$ are calculated by the Softmax function according to the relative magnitude of $L_{bo}$ and $L_{fo}$:

$$\omega_{bo} = e^{(1 - e^{L_{bo}}/(e^{L_{bo}} + e^{L_{fo}})) - 0.5} \quad (3)$$

The $\omega_{fo}$ can be computed in the same way. If $L_{bo}$ and $L_{fo}$ are closer, the weights $\omega_{bo}$ and $\omega_{fo}$ will be close to 1 and correspond to be a non-occluded region. On the contrary, the weights $\omega_{bo}$ and $\omega_{fo}$ will prone to be zero, and the occlusion is more likely to occur in one direction.

### B. Adaptive Cross Weighted Loss Between Optical Flow and Depth-pose

In this section, the loss function used for the self-supervised learning of camera pose, depth, and optical flow is described. These tasks are highly related by the scene re-projection. Assuming $p_t = [x, y]$ is a pixel in the target image $I_t$, then the re-projection coordinates $p_s'$ in the source views $I_s$ can be calculated by the rigid transformation $T_{s \to t}$:

$$[p_s', 1] = K T_{s \to t} D_t(p_t) K^{-1} [p_t, 1]^T \quad (4)$$

where $D_t$ is the depth predicted by the deep neural networks. The pixels $p_s'$ in the source images either belong to a rigid motion regions in (4) or a non-rigid motion regions. In this paper, inspired by the adaptive photometric loss in [9], we proposed an adaptive cross weighted loss of photometric and geometric error between the camera pose-depth and the optical flow to increase the amount of training data for each task and deal with the occlusions by the learned occlusion mask $M_{occ}$. The adaptive weights simultaneously consider the photometric error and geometric error between the cross tasks. For the photometric loss $L_o$ and $L_d$ of the optical flow and the depth and camera pose, we can compute the soft weight for each task by

$$\omega_o = [(1 - e^{L_o}/(e^{L_o} + e^{L_d})) > 0.28] \quad (5)$$

We find that just throwing away the bigger error during depth-pose and optical flow jointly training will achieve a better performance than (3). As the photometric loss is not reliable in the situations such as illumination variation, reflective surfaces, repetitive textures, here we further introduce the epipolar geometry to enhance the robustness of the model. The depth and pose estimation can reduce the rigid flow between adjacent views, thus the epipolar geometric loss between depth-pose and optical flow can also be uses to distinguish the motion regions.

To verify the displacements of the rigid flow and optical flow simultaneously by the geometric constraints and distinguish the moving objects, here we incorporate the adaptive weighted epipolar geometric constraints between the depth-pose and optical flow to adaptively mask out bad matches and non-rigid regions. The epipolar geometric constraints for optical flow and rigid flow based on the camera pose are $L_{Eo}$ and $L_{Ed}$ which can be calculated by

$$L_{Ed} = [p_s', 1] F [p_t, 1]^T \quad (6)$$

where $F = t_{s \to t} \times R_{s \to t}$ is the fundamental matrix. Thus the adaptive weights of geometric constraints for the rigid flow and the optical flow can be computed by (4). To improve the reliability of the above adaptive weighted loss, we use both the weights of photometric error and geometric error to constrain each loss. Thus, the cross weighted loss of the photometric and geometric error between the depth-pose and optical flow is

$$L_{ap}=\omega_{Ed}\omega_o(L_o+\lambda_{ed}L_{Ed})+\omega_{Eo}\omega_d(L_d+\lambda_{eo}L_{Eo}) \quad (7)$$

TABLE I.    QUANTITATIVE RESULTS COMPARED WITH THE STATE-OF-THE-ART SELF-SUPERVISED LEARNING METHODS FOR MONOCULAR DEPTH ESTIMATION ON THE KITTI DATASET [1] WITH EIGEN'S SPLITS AND DEPTH CAPPED AT 80M (WITHOUT POST-PROCESSING [6]).

| | Methods | Supervision | Error metrics | | | | Accuracy metrics | | |
|---|---|---|---|---|---|---|---|---|---|
| | | | *Abs Rel* | *Sq Rel* | *RMSE* | *RMSE log* | $\delta < 1.25$ | $\delta < 1.25^2$ | $\delta < 1.25^3$ |
| Trained on KITTI | Zhou et al. [7] | M | 0.198 | 1.836 | 6.565 | 0.275 | 0.718 | 0.901 | 0.960 |
| | Godard et al. [6] | S | 0.141 | 1.186 | 5.677 | 0.238 | 0.809 | 0.928 | 0.969 |
| | Vid2depth [35] | M | 0.159 | 1.231 | 5.912 | 0.243 | 0.784 | 0.923 | 0.970 |
| | Yin et al. [33] | M | 0.149 | 1.060 | 5.567 | 0.226 | 0.796 | 0.935 | 0.975 |
| | Shen et al. [44] | M | 0.139 | 0.964 | 5.309 | 0.215 | 0.818 | 0.941 | 0.977 |
| | Zhan et al. [30] | S | 0.135 | 1.132 | 5.585 | 0.229 | 0.820 | 0.933 | 0.971 |
| | Monodepth2 [47] | M | 0.115 | 0.903 | 4.863 | 0.193 | 0.877 | 0.959 | 0.981 |
| | Chen et al. [9] | M | 0.135 | 1.070 | 5.230 | 0.210 | 0.841 | 0.948 | 0.980 |
| | Xue et al. [37] | M | 0.113 | 0.864 | 4.812 | 0.191 | 0.877 | 0.960 | 0.981 |
| | Baseline | M | 0.120 | 0.928 | 5.034 | 0.196 | 0.852 | 0.951 | 0.977 |
| | Baseline+MaskFlow | M | 0.114 | 0.876 | 4.868 | 0.190 | 0.877 | 0.959 | 0.982 |
| | +cross weights | M | 0.112 | 0.825 | 4.752 | 0.188 | 0.878 | 0.960 | 0.982 |
| Trained on CS | GeoNet.[33] | M | 0.210 | 1.723 | 6.595 | 0.281 | 0.681 | 0.891 | 0.960 |
| | Casser et al. [45] | M | 0.153 | 1.109 | 5.557 | 0.227 | 0.796 | 0.934 | 0.975 |
| | GLNet. [9] | M | 0.129 | 1.044 | 5.361 | 0.212 | 0.843 | 0.938 | 0.976 |
| | Ours | M | 0.119 | 0.986 | 5.182 | 0.179 | 0.852 | 0.945 | 0.978 |

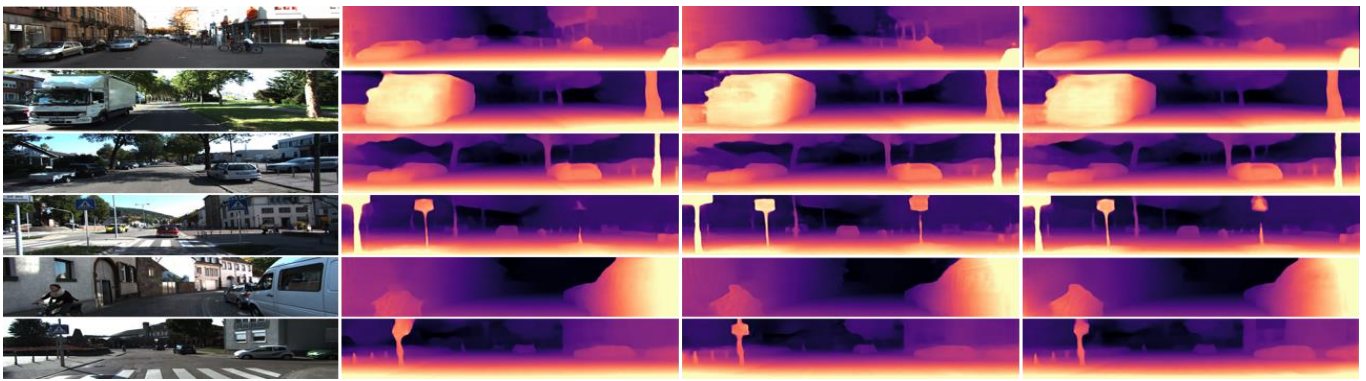

Fig. 2. Qualitative evaluation of depth estimation on the KITTI dataset [1]. The columns from left to right show respectively input images, the state-of-the-art predicted depth maps (Godard et al., 2019 [47]; Xue et al., 2020 [36]), and the depth maps obtained by the proposed method.

The reason behind this is that for a non-rigid regions, even if the pixel is correctly re-projected according to the camera's ego-motion, the photometric error would still be high but the geometric distance under the fundamental matrix $F$ obtained by the estimated camera pose would be proper. While the photometric error of the optical flow is proper for the non-rigid objects but the epipolar geometric loss based on the camera pose would be high. To ensure that such pixels are given low weights, we weigh them with their epipolar distance and photometric error compared with rigid flow and optical flow. Thus the total losses are a weighted sum over losses mentioned above, and sum over all valid image pixels.

$$L_{total}=L_{op}+L_{ap}+\lambda_s L_s \quad (8)$$

Where $L_s$ is the smoothness loss proposed in [47], $\lambda_s$ is the weight factor of this loss. These networks are trained jointly by means of self-supervised learning. Compared to previous work, our method could explicitly manage with occlusions, non-rigid motions and bad mismatches.

## IV. EXPERIMENTS

In this section, we conduct widely experiments on the performance evaluation and compare it with prior state-of-the-art approaches on multiple tasks.

### A. Implementation Details

**Datasets.** We mainly conduct extensive experiments on the benchmarking KITTI [1] and Cityscapes datasets [3], then validate the camera pose estimation on KITTI Odometry Dataset. We used the officially provided ground-truth labels for evaluation. We also trained the model on Cityscapes [3] and tested on KITTI [1]. The Make3D dataset [3] was used to demonstrate the generalization ability of the proposed method. For the raw KITTI dataset, the Eigen et al.'s split [8] was used for training and testing, similar to the related work [9],[47],[37]. All the input images were resized to 192×640, and the optical flow network was evaluated on KITTI 2015 [1] training dataset. For the KITTI odometry dataset, the sequences 00-08 was used for training and 09-10 for testing [39],[40]. Since the camera poses in the KITTI odometry dataset are relatively regular and steady, the original test sequences was also sampled to mimic the fast camera motions for testing on unseen challenging data.

**Parameters setting.** For all experiments, the weighting of the different loss components was set as $\lambda_{eo}$=0.02 and $\lambda_{ed}$=0.002 in (7), and $\lambda_s$=0.001 in (8). During training, batch normalization was used for all the layers except the output layers. We trained our model with the Adam optimizer with $\beta_1$=0.9, $\beta_2$=0.999, Gaussian random initialization, and a mini-batch size of 6. The learning rate was initially set to 0.0001 and halved it after every 10 epoch until the end. Additionally, random resizing, cropping, flipping and color jittering were used for data augmentation during training following [47]. The backbone network was initialized with ImageNet [22] weights, and optimized for maximum of 30 epochs.

TABLE II. VISUAL ODOMETRY RESULTS UNDER THE AVERAGE ABSOLUTE TRAJECTORY ERROR AND THE STANDARD DEVIATION IN METERS ON THE KITTI ODOMETRY DATASET [1].

| Methods | Sequence 09 | Sequence 10 | # frames |
|---|---|---|---|
| Zhou [7] | 0.021 ±0.017 | 0.020 ±0.015 | 5 |
| Mahjourian [35] | 0.013 ±0.010 | 0.012 ±0.011 | 3 |
| GeoNet [33] | 0.012 ±0.007 | 0.012 ±0.009 | 5 |
| Ranjan [46] | 0.012 ±0.007 | 0.012 ±0.008 | 5 |
| Monodepth2[47] | 0.017 ±0.008 | 0.015 ±0.010 | 2 |
| Shen et al. [44] | 0.009 ±0.005 | 0.008 ±0.007 | 3 |
| GLNet [9] | 0.011 ±0.006 | 0.011 ±0.009 | 3 |
| Ours | 0.008 ±0.005 | 0.008 ±0.006 | 3 |

TABLE III. THE OPTICAL FLOW ESTIMATION RESULTS ON KITTI 2015 TRAINING SET.

| Methods | Noc | All |
|---|---|---|
| UnFlow [31] | - | 8.10 |
| GeoNet [33] | 8.05 | 10.81 |
| DF-Net [30] | - | 8.98 |
| CC [46] | - | 5.66 |
| GLNet [9] | 4.86 | 8.35 |
| MaskFlowNet+$L_{op}$ | 4.35 | 8.13 |
| MaskflowNet+$L_{op}$+$L_{Eo}$ | 4.27 | 6.60 |

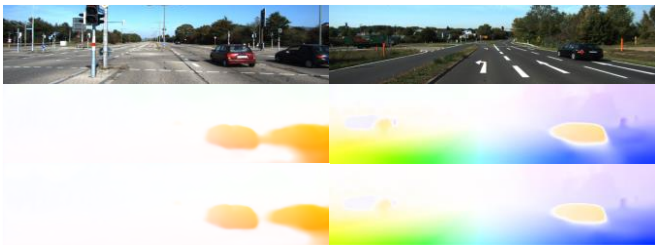

Fig. 3. Qualitative results of the predicted optical flow. Top Row: the Input images, Middle Row: MaskFlowNet trained by (2), Bottom Row: joint learning by (7).

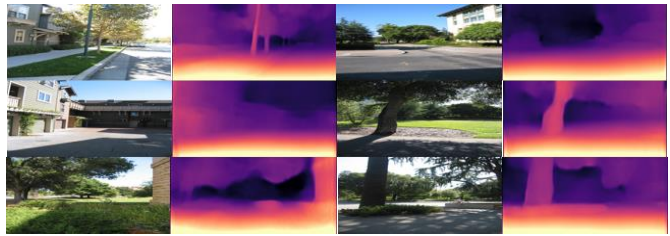

Fig. 4. Illustration of samples of the depth estimations on the Make3D dataset by the model only trained on the KITTI dataset.

**Network Architectures.** As our method focused on a better self-supervised learning scheme of monocular depth and camera pose, the similar depth and optical flow network structures were adopted in the existing methods [8],[47]. For the depth network, we used the same architecture as [47] which adopts ResNet18 [19] as encoder. The optical flow network was based on MaskFlowNet [9] and trained with the occlusion-aware photometric loss. The camera pose network was similar as in [47]. Firstly, we only trained the optical flow network in a self-supervised manner via occlusion aware photometric loss. After 10 epochs, we froze optical flow network and trained the depth and pose networks for another 10 epochs. Finally, we jointly trained all networks for last 10 epochs. In all experiments, the model in [9] and the minimum of the epipolar geometric loss among source views was used as the baseline.

### B. Evaluation of Depth Estimation

**Main Results on KITTI.** The evaluation of depth estimation follows previous works [9],[47]. For a fair comparison, we provided a comparison with the baseline, as well as the state-of-the-art methods. The measure criterion conformed to the one used in [47]. As shown in Table I, with the same underlying network structure, the proposed method achieved a better performance. Qualitative results could be seen in Fig. 2. It is shown that our method could reduce artifacts and improve the visual quality of the depth map. The proposed adaptive cross-weighted loss can significantly improve both the depth and the optical flow performance. While, the promotion for depth estimation is not as outstanding as optical flow estimation.

**Generalization ability on Make3D.** To illustrate that the proposed method was able to generalize to other datasets unseen, we used our network trained only on the KITTI dataset, despite the differences of the datasets both in contents and the camera parameters, the reasonable results still could be achieved. Qualitative results were shown in Fig. 4.

### C. Pose Evaluation

We further evaluated our method for visual odometry applications. We compared our method with other state-of-the-art depth-pose learning methods [7],[33],[9],[46] on the official KITTI odometry benchmark. We measured the Absolute Trajectory Error (ATE) over N-frame snippets (N=3 or 5), as measured in [47]. The estimation results in Table II showed the improvement over existing methods with the adaptive cross-weighted loss, we achieved certain performance improvements over state-of-the-art self-supervised learning methods.

### D. Optical Flow Evaluation

Similar to previous methods [9],[39], we used the training set for evaluation, as the optical flow in this paper was learned by self-supervised manner. Table III summarized the results of optical flow estimation using the average end-point-error (EPE) over non-occluded regions (Noc) and overall regions (All) on the KITTI 2015 training set. Results showed that MaskFlowNet [8] could also be well learned by the occlusion aware photometric loss and outperformed most previous self-supervised flow estimation methods [32],[31],[43]. Qualitative results were given in Fig. 3, showed that the flow quality was improved for rigidly and non-rigidly moving scene regions.

## V. CONCLUSIONS

We have presented an learnable occlusion aware optical flow-guided self-supervised learning framework to jointly learn depth, optical flow and camera pose from monocular videos. The model combines a learnable occlusion mask and novel cross weighted loss of photometric and geometric error between the cross tasks, optical flow and camera pose-depth, to solve the problems of occlusions and moving objects. The experiments show that the combination of the photometric loss and epipolar geometric constraints can achieve competitive performance.